\documentclass{article}
\usepackage{spconf,amsmath,epsfig}
\usepackage{tikz}
\usepackage{soul}
\usepackage{caption, subcaption}

\usetikzlibrary{calc}
\usepackage{soul}

\let\OLDthebibliography\thebibliography
\renewcommand\thebibliography[1]{
  \OLDthebibliography{#1}
  \setlength{\parskip}{0pt}
  \setlength{\itemsep}{0pt plus 0.3ex}
}

\begin{document}\sloppy

% Example definitions.
% --------------------
\def\x{{\mathbf x}}
\def\L{{\cal L}}

% Title.
% ------
\title{Realistic, Animatable Human Reconstructions for Virtual Fit-On}
%
% Single address.
% ---------------
\name{Gayal Kuruppu*\thanks{* Corresponding author. Email: gayalkuruppu.gk@gmail.com}, Bumuthu Dilshan, Shehan Samarasinghe, Nipuna Madhushan, Ranga Rodrigo}
% Address and e-mail should NOT be added in the submission paper. They should be present only in the camera-ready paper. 
\address{Department of Electronic and Telecommunication Engineering, University of Moratuwa, Sri Lanka}
% \toappear{Corresponding author. Email: gayalkuruppu.gk@gmail.com}
% \address{gayalkuruppu.gk@gmail.com, nipuna.madhushan96@gmail.com, bumuthu.dilshan@gmail.com, shehan.akhs@gmail.com, ranga@uom.lk}

                % \twoauthors{John Doe
                %       \thanks{This work was supported by...}}
                %             {Doe's address, department \\
                %             City, etc \\
                %             optional e-mail address}
                %             {Judy Smith}
                %             {Smith's address, department \\
                %             City, etc \\
                %             optional e-mail address}
\maketitle

\begin{abstract}

We present an end-to-end virtual try-on pipeline, that can fit different clothes on a personalized 3-D human model, reconstructed using a single RGB image. Our main idea is to construct an animatable 3-D human model and try-on different clothes in a 3-D virtual environment. The existing frame by frame volumetric reconstruction of 3-D human models are highly resource-demanding and do not allow clothes switching. Moreover, existing virtual fit-on systems also lack realism due to predominantly being 2-D or not using user's features in the reconstruction. These shortcomings are due to either the human body or clothing model being 2-D or not having the user's facial features in the dressed model. We solve these problems by manipulating a parametric representation of the 3-D human body model and stitching a head model reconstructed from the actual image. Fitting the 3-D clothing models on the parameterized human model is also adjustable to the body shape of the input image. Our reconstruction results, in comparison with recent existing work, are more visually-pleasing. 

% We use our own method for stitching the head model with the parameterized body model.
% Existing 3D volumetric methods that create human body models lack a way to change clothes in the model.

% Numerous results demonstrates the realisticness of our method compared to the existing recent work.

\end{abstract}
\begin{keywords}
virtual try on, 3-D virtual fit on, 3-D human reconstruction
\end{keywords}
\section{Introduction}
\label{sec:intro}
% \hl{1. Copy each paragraph to Word and check spelling and grammar. 2. Provide descriptive captions to each figure. Describe each figure in the text as well. There cannot be a figure or a table without an appropriate description in the text.}

%The online fashion industry has a huge market share in e-commerce. it keeps on growing and it will replace in-store shopping in near future. Unlike in other online markets, when it comes to fashion, some bridges need to be filled to match the online and offline user experiences. The biggest drawback of online fashion is that users have less idea about how the cloth will look on them. Fashion is rapidly changing and each product is different from one another which makes it hard for the customer to buy a clothing item without visualizing it. Surveys[] show high return rates for fashion items causing huge losses to both customers and the sellers. Customers like to get that feeling in front of a mirror in the store where they get to see the clothes on themselves for real.

Virtual clothes fitting receives much attention in the research community, as existing models still need improvements in the quality of reconstruction in general. Constructing the human model based on pose extraction is a frame-by-frame approach that suffers from poor quality and high computational power requirements. It does not facilitate fitting-on clothes and merely dresses the body with existing clothes in the frame. The general approach for virtual fitting is constructing a 3-D human model (e.g., parametric models such as SMPL model \cite{loper2015smpl}) followed by fitting-on clothes. This too needs improvement in the quality of the reconstruction, (e.g., using the head models and managing the seams). Methods that use volumetric representations of the body, instead of parameterized models, too are popular. However, existing volumetric models also suffer from the problem of clothes amounting to the volume and can still improve in quality.

In a general sense, the virtual fitting proceeds by predicting the size of the clothes and letting customers fit-on the clothing items to an avatar created on their own with given body measurements. This simple approach, although predicts the clothing sizes, does not give a convincing sensation to the user as the clothes are on an avatar, not a realistic reconstruction of the user's bodily features. Some studies use augmented reality to solve this problem with a digital mirror in a clothing store or a public place. These augmented reality solutions are mostly 2-D and have imperfections, such as misfitting and incorrect warping \cite{han2018viton}. In this context, 3-D body reconstruction and 3-D clothing fit-on approaches are much superior.

In this study, starting with body parameters extracted from the RGB image of the user, we manipulate an SMPL \cite{loper2015smpl} body with an articulated head model \cite{li2017learning} and fit-on clothes of choice to provide a more realistic, good-quality 3-D reconstruction which is also animatable. We present this virtual try-on system for online fashion with an in-store like user experience using only a single RGB image from the user. Through this, We create a unique and more refined animatable 3-D human body to realistically represent to the user. To the best of our knowledge, this is the first animatable 3-D clothes fit-on system with a realistic body model of the user with a personalized head.
% This model will be able to shop in the virtual store that we create to simulate the actual experience that we usually get in real life using 
The contributions of this paper are producing a personalized 3-D human model with a more realistic head model and producing an animatable human model that would be useful in AR-, VR-, or MR-based virtual try-on systems. In doing so, we have devised a pipeline for virtual try-on for fashion. We will make the code base available.

\section{Related Work}
Since 3-D modeling in computer vision has been rapidly developing in past few years, there are a significant number of recent works on virtual try-on. Nowadays, the virtual try-on goes beyond 2-D clothing transfer \cite{han2018viton, yang2020towards, raj2018swapnet, minar2020cp, jandial2020sievenet} with a much realistic user experience. 
We can recognize three major types of related works: 1) 2-D virtual try-on using image translation, 2) 3-D volumetric human model reconstruction, and 3) 3-D parametric virtual try-on systems. We do not discuss 2-D virtual try-on using image translation, as our work is based on 3-D human reconstruction \cite{saito2019pifu, saito2020pifuhd, alldieck2019learning, alldieck2019tex2shape, jackson20183d} and 3-D face reconstruction \cite{sanyal2019learning, deng2019accurate, zhu2017face, jackson2017large}. Methods for 3-D human reconstruction can be classified into two top-down approaches: 1) free-form, 2) model-based. The volumetric human model reconstruction is considered as the top-down free-form method. These methods are based on multi-view stereo reconstruction, and therefore require multiple RGB
or depth cameras. Meanwhile top-down model-based methods exploit a parametric body model consisting of pose and shape reconstruction.

In this paradigm, SMPL \cite{loper2015smpl} is an important model. SMPL framework enables 3D human reconstruction by parameterizing the human body shapes and the pose. Therefore, most of the recent works related to 3-D virtual fit-on are built on top of SMPL. The 3-D reconstruction based virtual fit-on systems needs major classes of techniques---3-D volumetric human model reconstruction and 3-D parametric virtual try-on systems---which we discuss in what follows.

\textbf{3D volumetric human model reconstruction:} Although the volumetric representation of clothed 3-D human model approach grows vastly, it cannot be considered as the ideal approach for a 3-D virtual try-on system. PIFu \cite{saito2019pifu} is a work that creates the 3D static body model based on SMPL from single or multiple images. It has proposed a pixel aligned implicit function representation for 3-D deep learning for the challenging problem of textured surface inference of clothed 3-D humans. However, PIFu generates a static model that is non-animatable and cannot be used for fitting garments. PIFuHD \cite{saito2020pifuhd} is an advancement of PIFu mainly focus on the accuracy and details of the 3-D model. Although the details have been improved, still the body model is stationary. 
Octopus \cite{alldieck2019learning} is a hybrid method combining bottom-up and top-down approaches. This work also creates a similar model to PIFu, reconstructing clothed human 3-D models. To build the 3-D model, an RGB video standing on in T pose would be used in Octopus. It synthesizes 3-D models using both bottom-up and top-down streams allowing information to flow in both directions. As the model is not parameterized but static, the model cannot be animated. In terms of virtual try-on, another major disadvantage is that it is not possible to make the unclothed mesh.

\textbf{Parametric virtual try-on systems:} 
We can see a lack of work on an end-to-end system that performs virtual try on with a fully-customized human model, perhaps due to the sub-problems being more technically engaging. Real-Time 3-D Model Reconstruction and Mapping for Fashion \cite{makarov2020real} introduce a method to specifically for virtual fit-on. This consists of a real-time animatable generic body model (SMPL model) enabling garment fitting on the mesh. This work uses the SMPL generic body model which has the same head though it has become animatable. This work can be differentiated by  the previous 3-D model reconstruction not only with animatability but also with the ability to change the clothes on it. However, this work still consists of the generic head that came from SMPL mesh. In terms of virtual try-on, this is a significant disadvantage for user experience.  Our approach overcomes this issue with an effective end-to-end virtual try-on system with a fully personalized parametric human 3-D mesh enabling fitting on different garment models.

\section{Methodology}

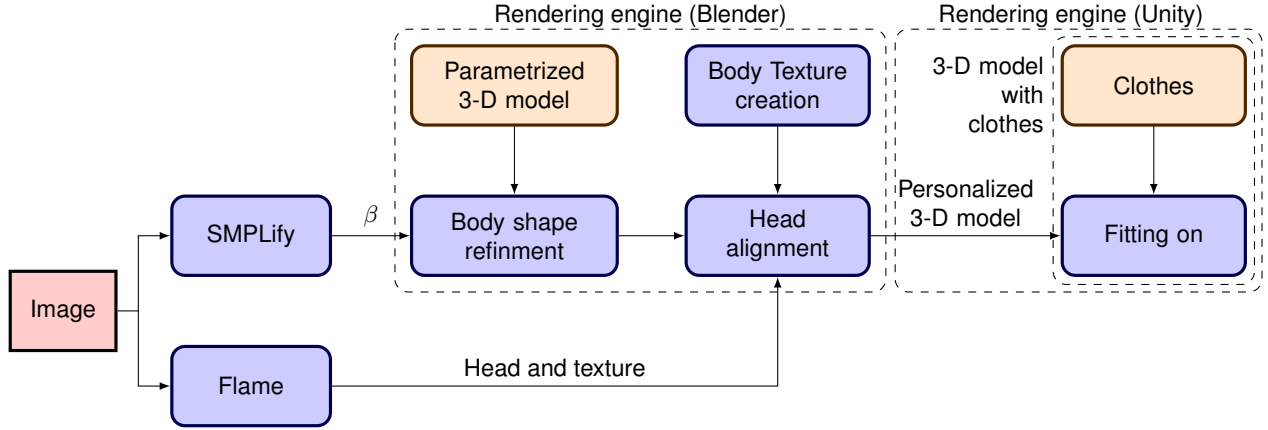
\begin{figure*}[h]
    \centering
    \begin{tikzpicture}[font=\sffamily\small]
\tikzset{node distance = 3.5cm and 1cm}
\tikzstyle{block} = [draw=blue!30!black, very thick, fill=blue!20, rectangle, rounded corners=5, minimum height=3em, minimum width=6em]
\tikzstyle{input} = [draw, very thick, fill=red!20, rectangle, minimum height=3em, minimum width=4em]
\tikzstyle{resource} = [draw=orange!30!black, very thick, fill=orange!20, rectangle, rounded corners=5, minimum height=3em, minimum width=6em]

\node [input, name=input] {Image};
\path (input) ++(1,0) coordinate  (branch);
\draw (branch) ++(1.5,1) node [block, name=simplify] {SMPLify};
\draw (branch) ++(1.5,-1) node [block, name=flame] {Flame};
\node  [block, right of=simplify, name=shaperef, text width = 2.5cm, text centered] {Body shape refinment};
\node  [block, right of=shaperef, name=headalign, text width = 2.2cm, text centered] {Head alignment};
\draw[latex-] (shaperef) -- ++ (0,2) node  [resource, name=param3d, text width = 2.5cm, text centered] {Parametrized 3-D model};
\draw[latex-]  (headalign) --++ (0,2) node  [block, name=bodytexture, text width = 2.2cm, text centered] {Body Texture creation};
\path (headalign) ++ (5,0) node  [block, name=fittingon, text width = 2.2cm, text centered] {Fitting on};
\draw[latex-] (fittingon) --++ (0,2) node  [resource, name=clothes, text width = 2.2cm, text centered] {Clothes};    
    
\draw[-latex] (input) -- (branch) |- (simplify);
\draw[-latex] (branch) |- (flame);
\draw[-latex] (simplify) -- (shaperef) node[midway, above] {$\beta$};
\draw[-latex] (shaperef) -- (headalign);   
\draw[-latex] (headalign) -- (fittingon) node[midway, above, text width=2cm, text centered] {Personalized 3-D model};
\draw[-latex] (flame) -| (headalign)  node[near start, above] {Head and texture};

\draw[dashed, rounded corners=5] ($(param3d.north west) + (-0.2, 0.2)$) rectangle ($(headalign.south east) + (0.2,-0.2)$);
\path (param3d.north west) -- (bodytexture.north east)  node[midway, above, yshift=0.1cm] {Rendering engine (Blender)};

\draw[dashed, rounded corners=5] ($(clothes.north west) + (-2.2, 0.2)$) rectangle ($(fittingon.south east) + (0.2,-0.2)$);
\path ($(clothes.north west) + (-2.2, 0)$) -- (clothes.north east)  node[midway, above, yshift=0.1cm] {Rendering engine (Unity)};

\draw[dashed, rounded corners=5] ($(clothes.north west) + (-0.1, 0.1)$) rectangle ($(fittingon.south east) + (0.1,-0.1)$);
\path ($(clothes.north west) + (-0.1, 0)$) -- ($(fittingon.south west) + (-0.1, 0)$)  node[near start, anchor=east, yshift=0.1cm, text width=1.5cm, align=right] {3-D model with clothes};

\end{tikzpicture}
    \caption{System block diagram: The system take an image of the user as the input. SMPLify \cite{bogo2016keep}  computes $\beta$ parameters, and FLAME extracts the head model.}
    \label{fi:system_block}
\end{figure*}

The goal of this work is to create an animatable 3-D human model from a single RGB image, and use that model as a mannequin in a virtual clothing store to try on different clothing. Our final human model is a function of shape parameters ($\beta$ parameters) from \cite{loper2015smpl} and head parameters from \cite{li2017learning}. We construct both the models separately and automatically stitch them
%using a Python script 
in Blender\footnote{https://www.blender.org/}. 
%The key observation we find in related work is the 
This solves the problem of the inability to personalize the head in the human model. Moreover, FLAME allows us to give a detailed and animatable head. We also create a virtual store in  Unity \footnote{https://unity.com/} and facilitate automated switching of clothes under a VR setting. 
We then map clothes models to 3D human body models using scale and position parameters. The trained model gives the scale and position parameters, taking the $\beta$ parameters as input.
Fig.~\ref{fi:system_block} shows the overall system block diagram.

\subsection{3D Human Model Construction}
\label{sec:3-D_human_model_cons}
We use the parametric human body representation, SMPL in reconstructing the body model, and the parametric human face presentation, FLAME,  in reconstructing the head along with texture. In this way, we get a more personalized 3-D human model that includes not only the shape of the body but also a personalized head instead of the generic one in SMPL. 
We use SMPLify to get the $\beta$ parameters for the SMPL model. 
% We use FLAME to get created the head along with its texture.

\textbf{Body shape representation: }
% SMPL theoretical insights, equations, so on
We represent the shape of the human body model using the SMPL parametric representation, which represents the undressed body. The body model is defined as an explicit function of shape $\beta$, pose $\theta$, and translation $\gamma$. The function returns a triangulated surface with 6890 vertices. Shape parameters $\beta$ are coefficients of low-dimensional shape space, obtained using principal components after training on thousands of body scans. We use SMPLify to extract $\beta$ parameters from the image and use these $\beta$ parameters to reconstruct the body model. 

\textbf{Face representation: }
% FLAME theoretical insights, equations, so on
% For representing the face too, we use a parametric representation. 
We use the model formulation of FLAME to obtain a parametric representation for the face too. It consists of 5023 vertices. A function of shape $\beta$, pose $\theta$, and expression $\psi$ describe the face. We explain more about the combination of the head model and the body model in the Sec.~\ref{sec:combination}.

\textbf{Texture Generation: }
Texture Generation is a crucial part of the final outcome of the 3-D human model. The texture generated by FLAME is of good quality, and hence we use it as the texture of the head of the 3-D body model. Then we analyze the skin color of the user by the skin present in the user's head, as the head usually reveals the skin color of most of the users. Then we find the dominant skin color of the user and create the texture of the 3-D model's body. But, the texture we get for the body can be different from the texture of the head. Therefore we apply the texture of the body to the areas of the neck too to combine the textures without a seam.

\subsection{Combining the Head and the Body}
\label{sec:combination}

First, we cut the body model from the neck using the automated python script in Blender. Unlike a pre-designed 3-D model, we have to deal with the head model which is created for the given input image of the user. The advantage of the FLAME head model is that it has common ids for vertices in Blender. Therefore we have defined the group of vertices in the 3-D head model which has to be cut. As we mentioned earlier, SMPL is a parameterized human body model, and we have to add shape parameters that are generated according to the input image of the user. We read the generated shape parameters this stage and add them into the SMPL body model in Blender. The crucial task in the process of combining two models is placing the head model on the top of the neck of the body model. To do that, 1) rotate (in $y$- and $z$-direction) the head model to align with the body model, and 2) align the bottom of the head model with the top of the neck of the body model. 

\begin{figure}[h]
\begin{minipage}[b]{0.24\linewidth}
  \includegraphics[width=2cm, height=2.2cm]{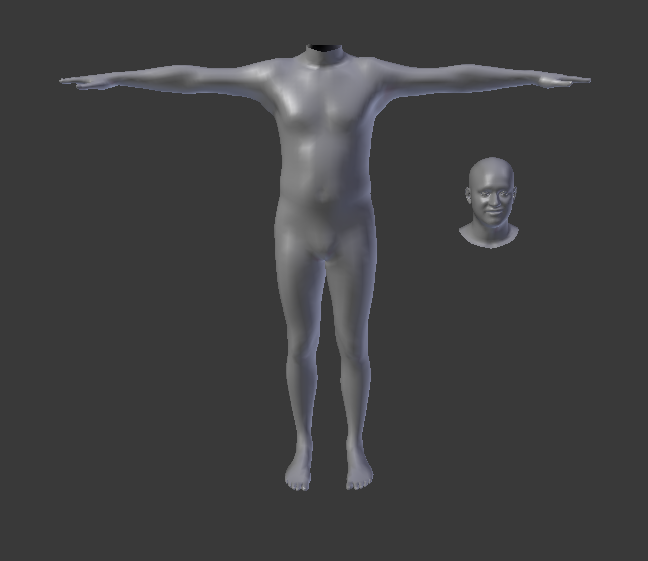}
  \centering
%  \vspace{1.5cm}
  \centerline{(a)}\medskip
\end{minipage}
\hfill
\begin{minipage}[b]{0.24\linewidth}
  \includegraphics[width=2cm, height=2.2cm]{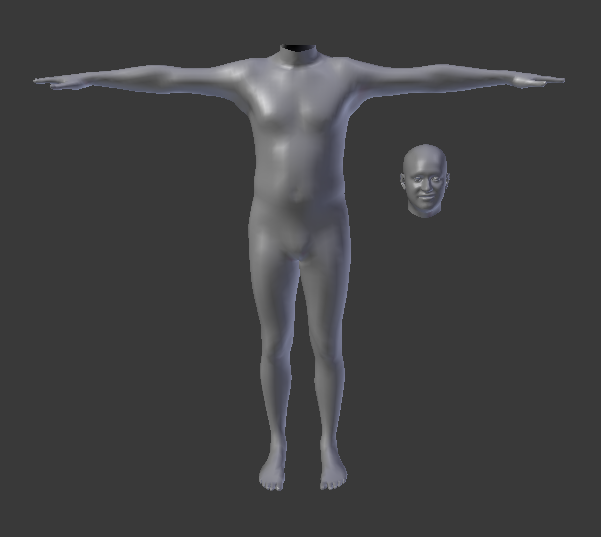}
  \centering
%  \vspace{1.5cm}
  \centerline{(b)}\medskip
\end{minipage}
\hfill
\begin{minipage}[b]{0.24\linewidth}
  \includegraphics[width=2cm, height=2.2cm]{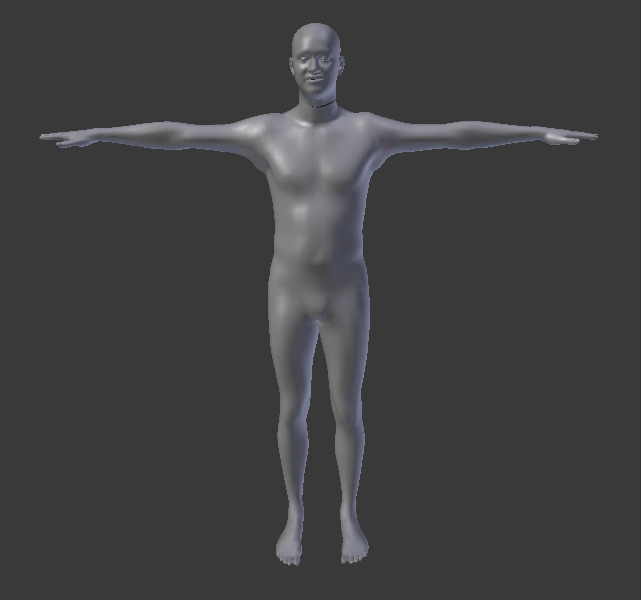}
  \centering
%  \vspace{1.5cm}
  \centerline{(c)}\medskip
\end{minipage}
\hfill
\begin{minipage}[b]{0.24\linewidth}
  \includegraphics[width=2cm, height=2.2cm]{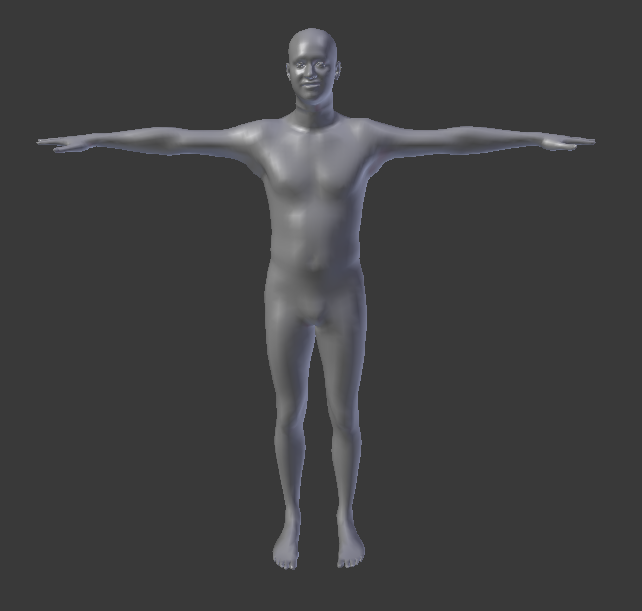}
  \centering
%  \vspace{1.5cm}
  \centerline{(d)}\medskip
\end{minipage}
\caption{3-D body model construction: (a) SMPL body model and FLAME head model, (b) Head model is cut from the neck, (c) Head model is rotated and moved to the top of the neck, (d) Both body and head models are combined}
\label{fig:res}
\end{figure}

After placing the head model on top of the body model, the head model is rigged to the bone structure of SMPL. Then the two models are combined by filling the space between the models. Fig.~\ref{fig:res} shows the process of stitching the head and generating the full 3-D body model.
% At the end of the process, the personalized 3D model is generated as an .fbx file which can be animated.

\begin{figure}[h]
\begin{minipage}[b]{0.48\linewidth}
  \includegraphics[width=4cm, height=4.4cm]{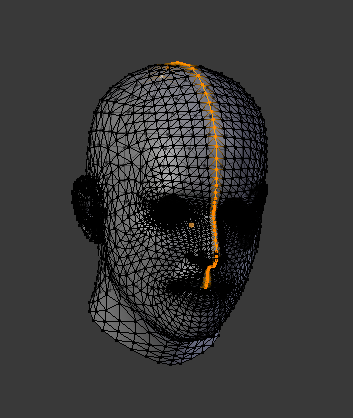}
  \centering
%  \vspace{1.5cm}
  \centerline{(a)}\medskip
  \label{fi:3-d_head_mesh-a}
\end{minipage}
\hfill
\begin{minipage}[b]{0.48\linewidth}
  \includegraphics[width=4cm, height=4.4cm]{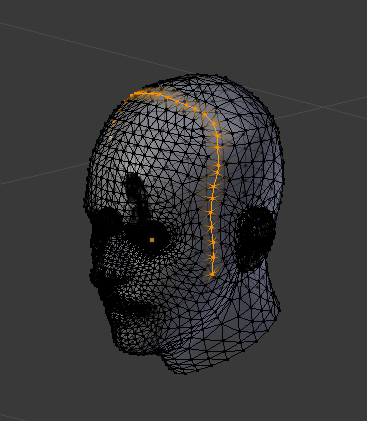}
  \centering
%  \vspace{1.5cm}
  \centerline{(b)}\medskip
  \label{fi:3-d_head_mesh-b}
\end{minipage}
\caption{3-D mesh of the head model: (a) Selected set of vertices along the symmetric line, (b) Selected set of vertices in the direction of $z$-axis.}
\label{fi:3-d_head_mesh}
\end{figure}

\textbf{Head and Body Alignment}
We combine the head model and the body model in a 3-D space. For this, we need to align the head model at the top of the body model. We rotate and move the head model in order to align with the body model. Fig.~\ref{fi:3-d_head_mesh} (a) shows that the 3-D mesh of the head model is symmetric along the $x$-axis. We select a set of vertices along the symmetric line of the 3-D mesh of the head model. Eq.~\ref{eq:rot_angle1} describes the error function ($E_y$) for the rotational angle along $y$-axis by using the set of vertices ($V$) selected where, $\alpha$ and $\sigma$ are the mean and the variance of $z$ coordinates in $V$. The error function is calculated based on the mean error of $x$ coordinate ($|{P_i}_x-\overline{P_x}|$) of the set of vertices selected. We have considered an exponential weighted distribution of the $z$ coordinates ($P_z$) of the set of vertices as a weighted combination of points. The error function ($E_z$) for $z$-axis in Eq.~\ref{eq:rot_angle2} is very similar to the $E_y$ but the weight distribution is different. In the Eq.~\ref{eq:rot_angle3}, we calculate the rotational angle (${R_y}_i$) along the $y$ axis which is the same formula for the rotational angles along the $z$-axis and $x$-axis as well. $C_y$ is a variable that depends on the difference between the previous error and the current error of rotation, and it is used to vary the speed of the rotation and the direction (clockwise or anti-clockwise). We do the calculations for several iterations until the head rotates to its rightful position. The rotations along the $y$-axis and $z$-axis iterates parallel because of the symmetry.

%Eqs. \ref{eq:rot_angle1} through \ref{eq:rot_angle6} describe the process.

\begin{equation}\label{eq:rot_angle1}
E_y = \sum_{P_i \epsilon V} \ |{P_i}_x-\overline{P_x}| \ \exp{\left( \frac{{P_i}_z-\alpha}{\sigma} \right) ^2}
\end{equation}
\begin{equation}\label{eq:rot_angle2}
E_z = \sum_{P_i \epsilon V} \ |{P_i}_x-\overline{P_x}| \ \left( 1 - \exp{\left( \frac{{P_i}_z-\alpha}{\sigma} \right) ^2} \right)
\end{equation}
\begin{equation}\label{eq:rot_angle3}
{{R_y}_i} = {R_y}_{i-1} + 360 \times C_y \times E_y
\end{equation}
%\begin{equation}\label{eq:rot_angle4}
%(R_z)_i = (R_z)_{i-1} + 360 \times C_z \times E_z
%\end{equation}
We do the rotation along the $x$-axis as well by using a new set of vertices as shown in the Fig.~\ref{fi:3-d_head_mesh} (b). Eq.~\ref{eq:rot_angle5} shows the error function for the rotation of the $x$-axis where, $d$ is a learning parameter and $\gamma$ is the mean of the $z$ coordinates.

\begin{equation}\label{eq:rot_angle5}
E_x = \sum_{P_i \epsilon V} \ |{P_i}_y - d| \ \exp{\left( \frac{{P_i}_z-\gamma}{\sigma} \right) ^2}
\end{equation}
%\begin{equation}\label{eq:rot_angle6}
%(R_x)_i = (R_x)_{i-1} + 360 \times C_x \times E_x
%\end{equation}

\subsection{Clothes Fitting}

For the outcome to look natural, we must properly model the clothes. We create 3-D models of clothes by using the cloth simulation in Blender. We use cloth simulation to build 3-D clothes using 2-D block patterns. To complete the automated virtual try-on pipeline, we use pre-designed 3-D models for every cloth and query them accordingly.
%The main difficulty in the first method was that we need to have very high computational resources to create 3D models for garment items in real-time. Therefore we use the second method for cloth modeling. Since the first method needs high computational power, we choose the second method for cloth modeling.
The 3-D model of the clothing item may overlap with the body model in the virtual dressing room. To overcome that issue, we need to apply cloth physics (i.e., gravity, collisions, wind, stiffness) to each cloth. When we apply the cloth physics for different clothes with a personalized human body model, we align the 3-D model of the cloth with the personalized human body model.

% cloth fitting prediction starts
%Since the height and other body parameters of the personalized 3-D human model depends on the user, we need to align the garment accordingly.
%We have trained a regression model for the XYZ coordinates of the 3D model of garment items according to the shape parameters of the personalized 3D human body model. We have used image datasets[REF] of people to train the model.

%\begin{table}[h]
%\begin{center}
%\caption{Data Format for the model} \label{tab:cap}
%\begin{tabular}{|p{0.2cm}|p{0.2cm}|p{0.2cm}|p{0.2cm}|p{0.2cm}|p{0.2cm}|p{0.2cm}|p{0.2cm}|p{0.3cm}|p{0.%3cm}|p{0.3cm}|p{0.3cm}|p{0.3cm}|}
%  \hline
%  \multicolumn{10}{|c|}{Beta Parameters} & \multicolumn{3}{|c|}{Scale}\\
%  \hline
%  % after \\: \hline or \cline{col1-col2} \cline{col3-col4} ...
%  	\scriptsize B0 & 	\scriptsize B1 & 	\scriptsize B2 & 	\scriptsize B3 & \scriptsize B4 & %\scriptsize B5 & \scriptsize B6 & \scriptsize B7 & \scriptsize B8 & \scriptsize B9 & \scriptsize x & %\scriptsize y & \scriptsize z
%  \\
%  \hline
%\end{tabular}
%\end{center}
%\end{table}

\section{Experimental Results}
In summary, with the personalized 3-D body and head model, texture mapping and fitting-on clothes, we have an animatable realistic-looking virtual fit-on system. We qualitatively evaluate the results of our approach on images from three different online fashion store websites: Zalando \footnote{https://www.zalando.co.uk/women-home/}, Tom Tailor \footnote{https://www.tom-tailor.eu/}, and SSense \footnote{https://www.ssense.com/}. %We  present our 3-D human body model results for the dataset synthesised of human images (Sec.~\ref{ss:res-3d-human-body}) from websites \cite{website1, website2, website3, website4}. 
%We present our 3-D human body model construction results as well as the cloth fitting. 

\subsection{3D Human Body Model}
\label{ss:res-3d-human-body}
\begin{figure*}[h!]
  \centering
  %row 1
  \begin{subfigure}[b]{0.12\linewidth}
    \includegraphics[width=\linewidth]{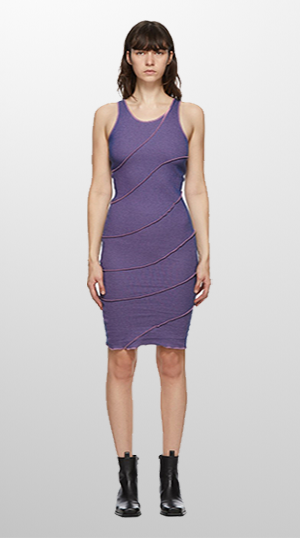}
%    \caption{}
  \end{subfigure}
 \begin{subfigure}[b]{0.12\linewidth}
    \includegraphics[width=\linewidth]{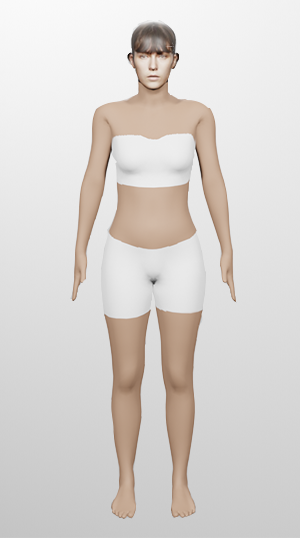}
%    \caption{}
  \end{subfigure}
  \begin{subfigure}[b]{0.24\linewidth}
    \includegraphics[width=\linewidth]{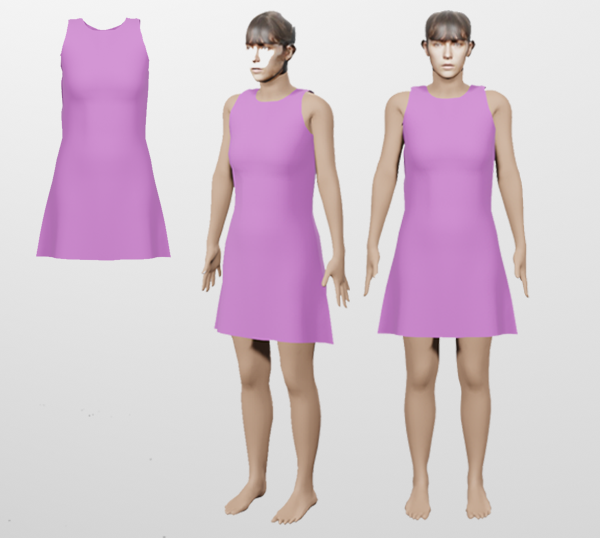}
%    \caption{}
  \end{subfigure}
  \begin{subfigure}[b]{0.24\linewidth}
    \includegraphics[width=\linewidth]{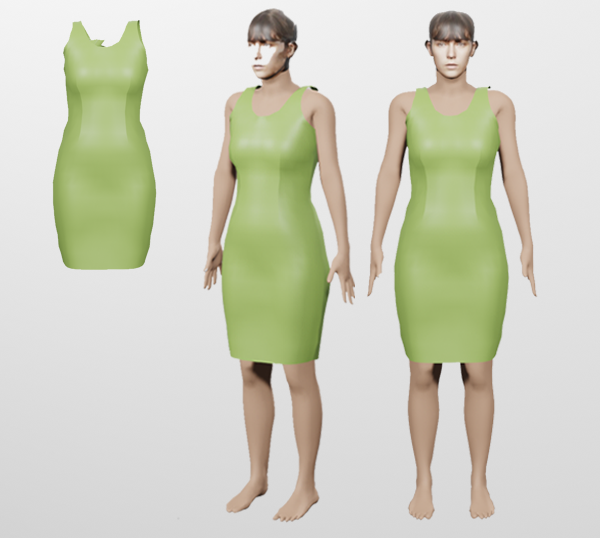}
%    \caption{}
  \end{subfigure}
  \begin{subfigure}[b]{0.24\linewidth}
    \includegraphics[width=\linewidth]{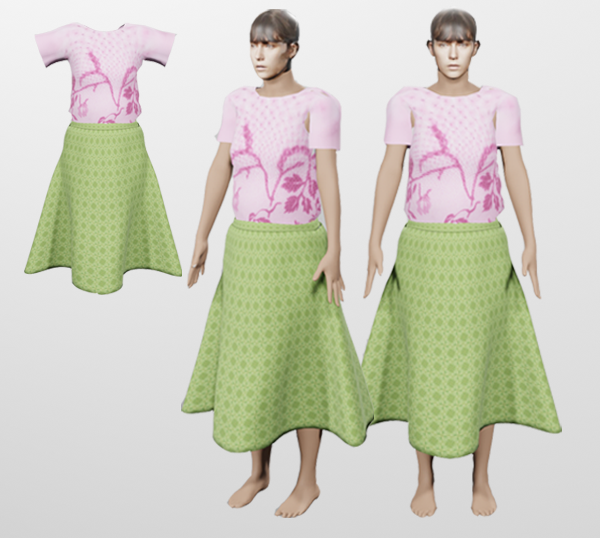}
%    \caption{}
  \end{subfigure}
  
  %row 2
  \begin{subfigure}[b]{0.12\linewidth}
    \includegraphics[width=\linewidth]{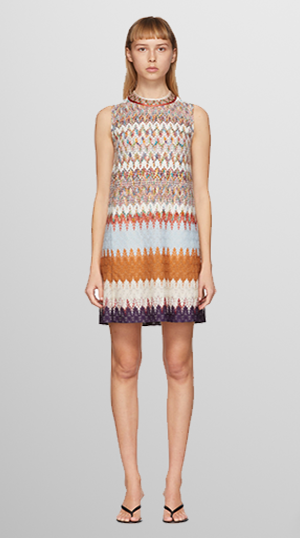}
    % \caption{}
  \end{subfigure}
 \begin{subfigure}[b]{0.12\linewidth}
    \includegraphics[width=\linewidth]{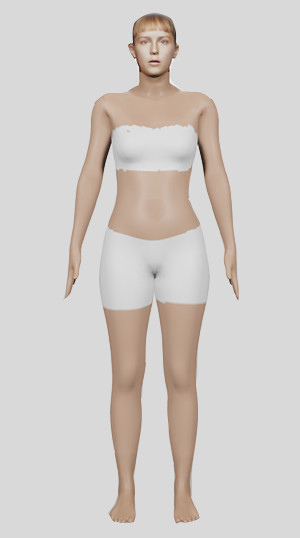}
    % \caption{}
  \end{subfigure}
  \begin{subfigure}[b]{0.24\linewidth}
    \includegraphics[width=\linewidth]{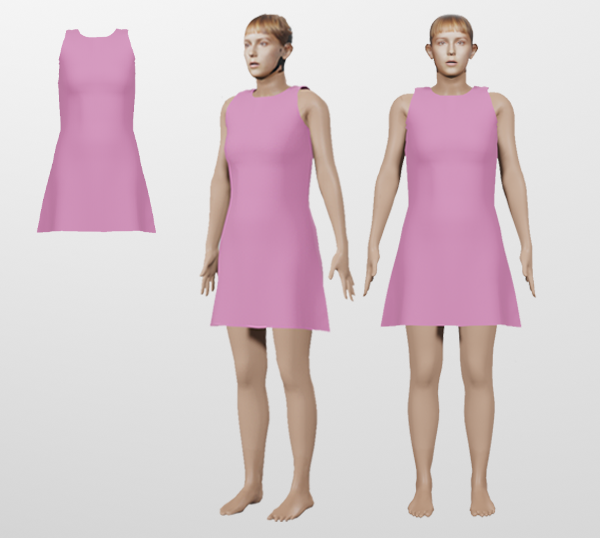}
    % \caption{}
  \end{subfigure}
  \begin{subfigure}[b]{0.24\linewidth}
    \includegraphics[width=\linewidth]{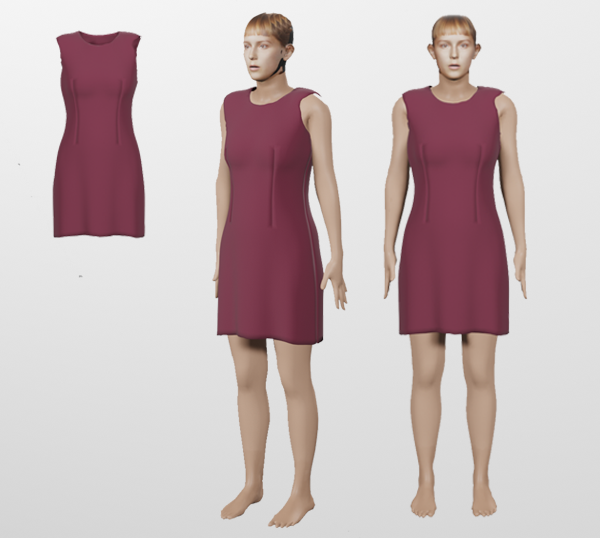}
    % \caption{}
  \end{subfigure}
  \begin{subfigure}[b]{0.24\linewidth}
    \includegraphics[width=\linewidth]{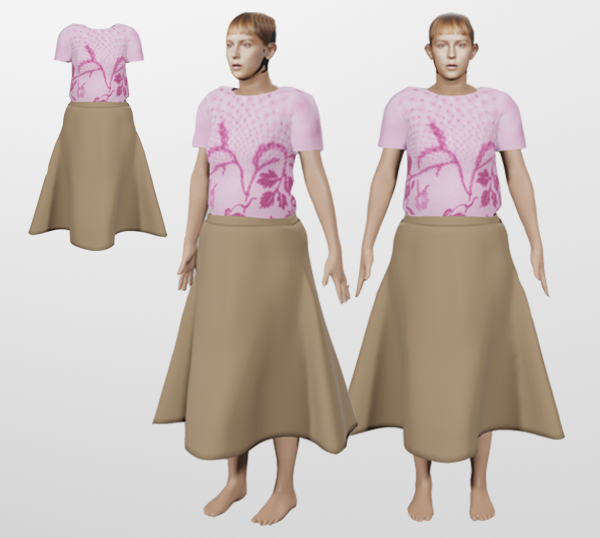}
    % \caption{}
  \end{subfigure}
  
  % row 3
  \begin{subfigure}[b]{0.12\linewidth}
    \includegraphics[width=\linewidth]{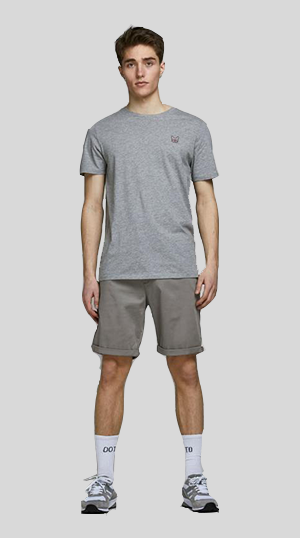}
    \caption{}
  \end{subfigure}
 \begin{subfigure}[b]{0.12\linewidth}
    \includegraphics[width=\linewidth]{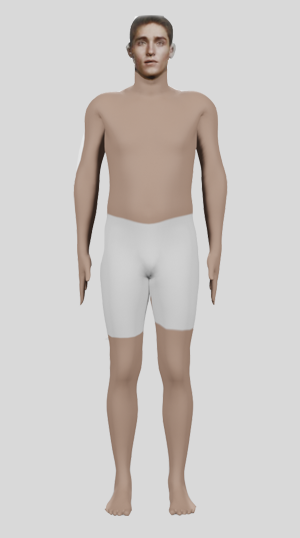}
    \caption{}
  \end{subfigure}
  \begin{subfigure}[b]{0.24\linewidth}
    \includegraphics[width=\linewidth]{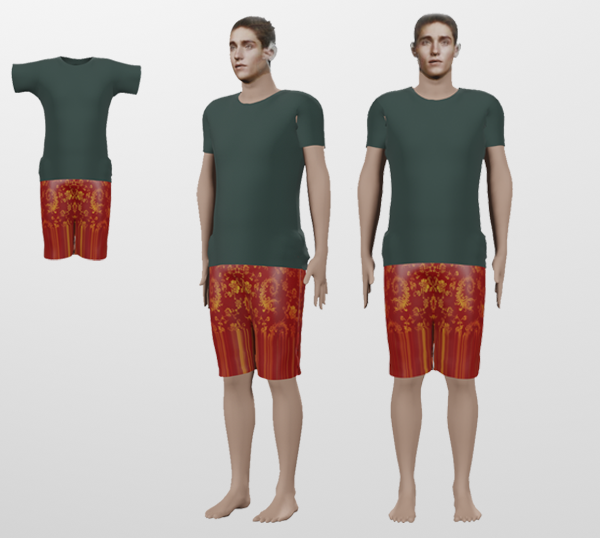}
    \caption{}
  \end{subfigure}
  \begin{subfigure}[b]{0.24\linewidth}
    \includegraphics[width=\linewidth]{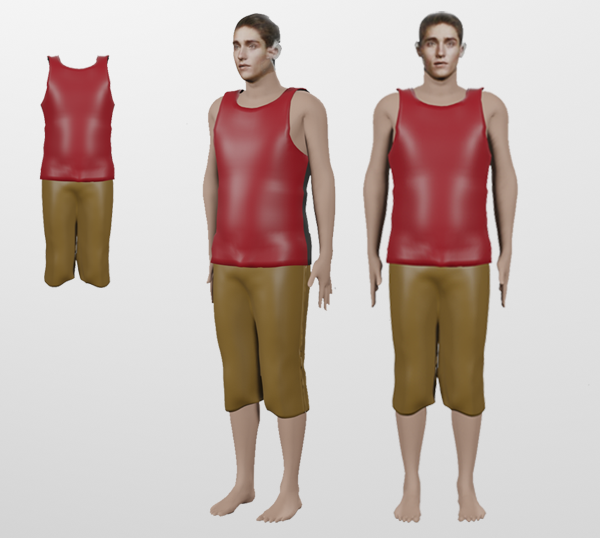}
    \caption{}
  \end{subfigure}
  \begin{subfigure}[b]{0.24\linewidth}
    \includegraphics[width=\linewidth]{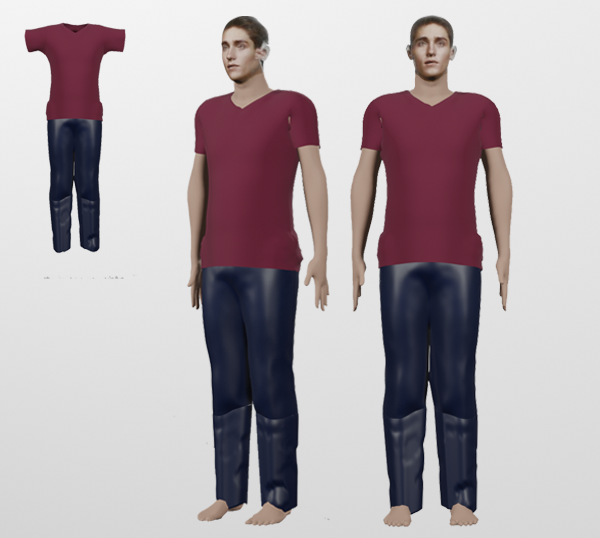}
    \caption{}
  \end{subfigure}
 
  \caption{\textbf{Results} on clothes switching with images of three different online fashion store websites (See text for sources). We have added the texture of the human body model as well as the 3-D models of clothes fitted to the body model: (a) Input image, (b) Unclothed personalized 3-D model reconstructed using our method, (c), (d), and (e) Clothes fitting and switching on the recreated personalized model.}
\label{fig:3d_model_cons}
\end{figure*}

We obtained the reconstitutions from RGB images of different humans, and fit-on different clothes on the personalized human model created with our method. Notice that the head is personalized, and the head-body seam is finely stitched.
As we can see in Fig.~\ref{fig:3d_model_cons}, our end-to-end approach of virtual try-on enables cloth switching on the fully personalized body. 

% Overall Results

% Different Poses
\begin{figure}[h]
  \centering
  \begin{subfigure}[b]{0.24\linewidth}
    \includegraphics[width=\linewidth]{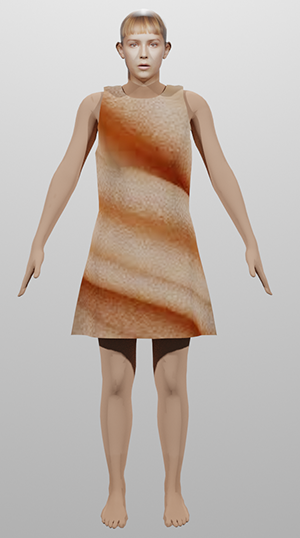}
%    \caption{}
  \end{subfigure}
 \begin{subfigure}[b]{0.24\linewidth}
    \includegraphics[width=\linewidth]{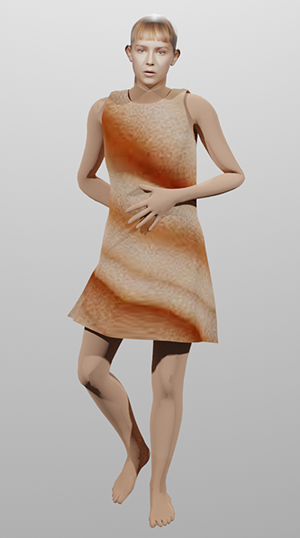}
%    \caption{}
  \end{subfigure}
  \begin{subfigure}[b]{0.24\linewidth}
    \includegraphics[width=\linewidth]{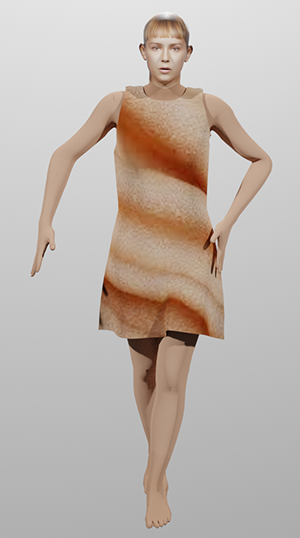}
%    \caption{}
  \end{subfigure}
  \begin{subfigure}[b]{0.24\linewidth}
    \includegraphics[width=\linewidth]{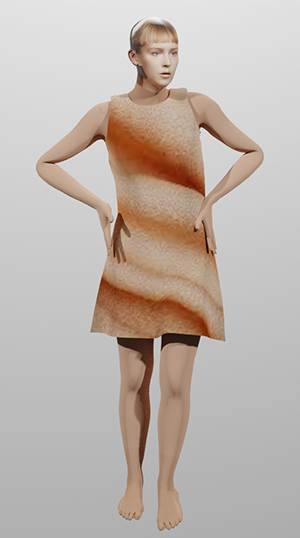}
%    \caption{}
  \end{subfigure}
  \caption{Different poses with the animatable 3-D human model including the clothes. The clothes move according to the pose of the human body model and behave in a realistic manner with physics.}
  \label{fig:pose}
\end{figure}

% Comparison with PiFu
\begin{figure}[h]
% row 1
  \centering
  \begin{subfigure}[b]{0.24\linewidth}
    \includegraphics[width=\linewidth]{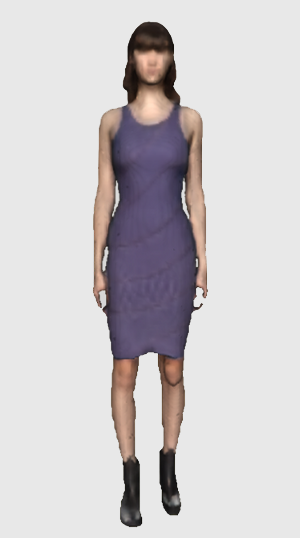}
%    \caption{}
  \end{subfigure}
 \begin{subfigure}[b]{0.24\linewidth}
    \includegraphics[width=\linewidth]{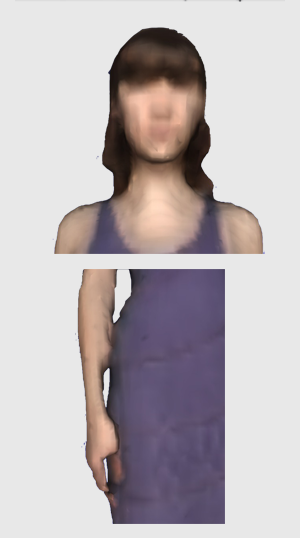}
%    \caption{}
  \end{subfigure}
  \begin{subfigure}[b]{0.24\linewidth}
    \includegraphics[width=\linewidth]{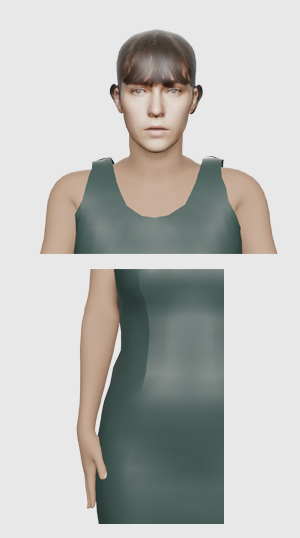}
%    \caption{}
  \end{subfigure}
  \begin{subfigure}[b]{0.24\linewidth}
    \includegraphics[width=\linewidth]{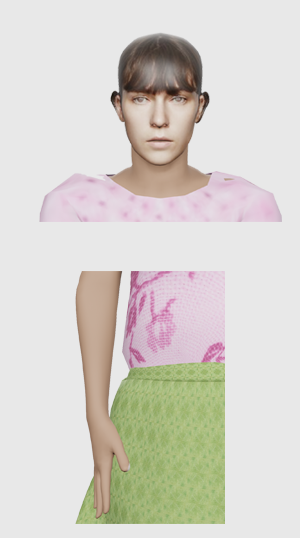}
%    \caption{}
  \end{subfigure}
  
  % row 2
  \begin{subfigure}[b]{0.24\linewidth}
    \includegraphics[width=\linewidth]{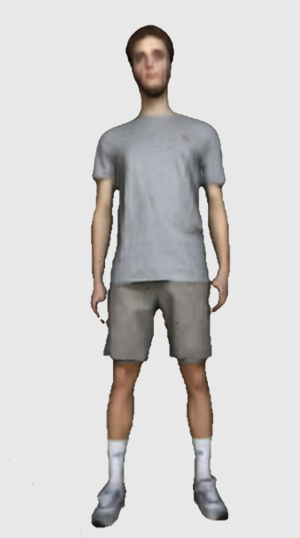}
    \caption{}
  \end{subfigure}
 \begin{subfigure}[b]{0.24\linewidth}
    \includegraphics[width=\linewidth]{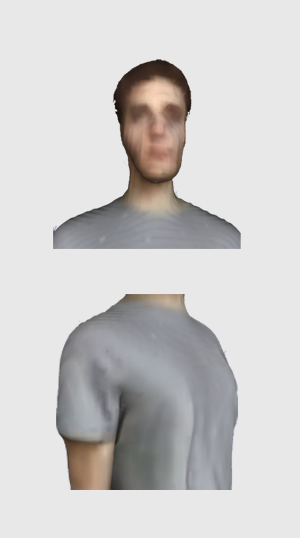}
    \caption{}
  \end{subfigure}
  \begin{subfigure}[b]{0.24\linewidth}
    \includegraphics[width=\linewidth]{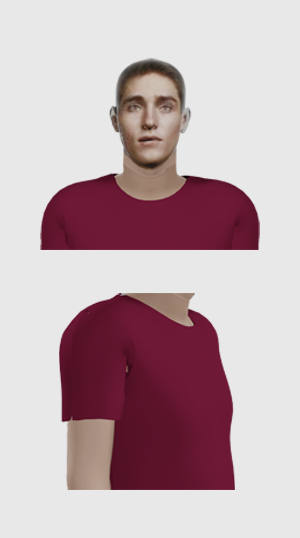}
    \caption{}
  \end{subfigure}
  \begin{subfigure}[b]{0.24\linewidth}
    \includegraphics[width=\linewidth]{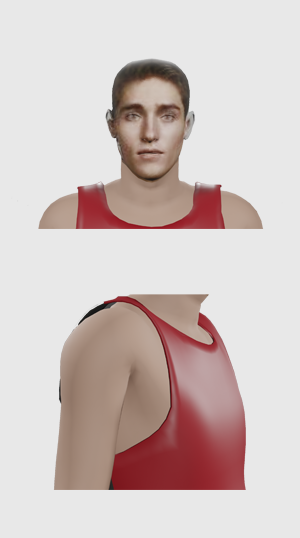}
    \caption{}
  \end{subfigure}
  \caption{\textbf{Result comparison with volumetric representations} (a) 3-D mesh created by PIFu, (b) Quality issues, (c) and (d) How our method overcome those issues. We used PIFu representing volumetric reconstruction in virtual try on. In (b), it shows how the low quality result look with issues in output quality, while (c) and (d) show the high quality output and ability to try on different clothes as well.}
  \label{fig:comp_pifu}
\end{figure}

\subsection{Clothes Fitting}
\label{ss:res-cloth-fitting}
Our virtual try-on environment is capable of switching the clothes according to the user's preferences. Fig.~\ref{fig:3d_model_cons} shows results of cloth fitting of different 3-D models of clothes on the same personalized 3-D human body model. The efficiency of rendering software is very essential for this task and we use Unity. When it comes to the 2-D mapping of clothes to a body model, we rarely see natural warping of a cloth which maps to the pose of the body model in the literature. In our work, the cloth warps naturally with the movements and poses in a realistic manner. Fig.~\ref{fig:pose} shows the behavior of cloth for different poses. 
% We fit the 3-D models of clothes in 13 different categories which have been defined in the DeepFashion2 \cite{DeepFashion2} dataset and use the basic block patterns for each category.

\subsection{Comparison with other work}
% PIFu is a deep learning-based clothed 3D human reconstruction work where we cannot customize the garments.
In this section, we qualitatively compare the differences between PIFu and our method. As Fig.~\ref{fig:comp_pifu} shows, we can see a clear improvement of the quality of the results compared to PIFu which uses a volumetric method for reconstructing the human 3-D models. The model given by that method gives imperfections as Fig.~\ref{fig:comp_pifu} (b) denotes. On the other hand, we can see that it is not able to use in changing clothes dynamically without recreating a different model for the next frame. Therefore, our method outperforms the existing works not only in the quality of the 3-D human model but also in the ability to virtually try on efficiently. 

% \clearpage

\section{Conclusion}
We have proposed a simple, yet efficient, end-to-end pipeline to reconstruct a personalized human model enabling clothes fitting as well as clothes switching for virtual try-on. The key idea of this work is to create a one-time personalized model without recreating the model frame by frame. Therefore, it enables clothes fitting, clothes switching, and animatability of the same 3-D model which are essential for the virtual try-on pipeline. Using this method, we have overcome the presence of imperfections that occurs in related volumetric approaches. 
% In our method, we reconstruct the 3-D human model using the parameterized framework and simultaneously generate the head separately. Then we combine them together and create our new model which is ready for virtual try-on. 

One limitation of our methods is the back-side of the head not being texture-mapped. We also plan to integrate hair models with our existing model in the future.

% In virtual try-on systems, size prediction is an important feature. We will be able to extend this work by predicting the cloth sizes.


\bibliographystyle{IEEEbib}
\bibliography{references}

\end{document}